\documentclass[runningheads]{llncs}
\usepackage[T1]{fontenc}
\usepackage{booktabs}
\usepackage{multirow}
\usepackage{graphicx}
\usepackage{amsmath}
\usepackage[table,xcdraw]{xcolor}
\usepackage{bbm}
\usepackage{hyperref}
\usepackage{amssymb}
\usepackage{float}
\usepackage{caption}
\usepackage{subcaption}
\usepackage{marvosym}
\usepackage{ifsym}
\usepackage{makecell}
\hypersetup{hypertex=true,
colorlinks=true,
linkcolor=blue,
anchorcolor=blue,
citecolor=blue}
\begin{document}
    \title{SCANet: Split Coordinate Attention Network for Building Footprint Extraction}
    \titlerunning{SCANet: Split Coordinate Attention Network} 
    
    \author
    {
        Chunshi Wang   
        \and
        Bin Zhao     \textsuperscript{(\Letter)} 
        \and
        Shuxue Ding  \textsuperscript{(\Letter)} 
    }
    
    \authorrunning{C. Wang and B. Zhao et al.}
    
    \institute{
    School of Artificial Intelligence, Guilin University of Electronic Technology
    \\
    \email{zhaobinnku@mail.nankai.edu.cn, sding@guet.edu.cn} 
    }
\maketitle              
\begin{abstract}
Building footprint extraction holds immense significance in remote sensing image analysis and has great value in urban planning, land use, environmental protection and disaster assessment. Despite the progress made by conventional and deep learning approaches in this field, they continue to encounter significant challenges.  This paper introduces  a novel plug-and-play attention module, \textbf{Split Coordinate Attention (SCA)}, which ingeniously captures spatially remote interactions by employing two spatial range of pooling kernels, strategically encoding each channel along x and y planes, and separately performs a series of split operations for each feature group, thus enabling more efficient semantic feature extraction. By inserting into a 2D CNN to form an effective SCANet, our SCANet outperforms recent SOTA methods on the public Wuhan University (WHU) Building Dataset and Massachusetts Building Dataset in terms of various metrics.  Particularly SCANet achieves the best IoU, 91.61\% and 75.49\% for the two datasets. Our code is available at \url{https://github.com/AiEson/SCANet}.

\end{abstract}
\section{Introduction}
Building footprint extraction on remote sensing images provides valuable guidance for urban planning, land use, environmental protection and disaster assessment. 
However, some factors may influence the acquisition of remote sensing images, such as sensors, meteorology and topography, leading to their variety quality and resolution, which poses a challenge for building footprint extraction.

Many traditional methods had been proposed for building footprint segmentation in the early years. 
For instance, Cui et al.  performed building footprint extraction by geometric elements of buildings, including line segments, edges and corner points \cite{trad2012}. 
In addition, some methods extracted buildings footprint by texture features, spectral information and LiDAR data of buildings \cite{guangpu2011per,lidar_and_texture,lidar1,lidar0}.
However, the above-mentioned methods require a mass  of handy-crafted features and have strong dependence on the quality and resolution of remote sensing images, which inevitably  lead to  poor generalization capacity over different remote sensing images.

Recently, many deep learning based methods have made significant progress in image processing \cite{pspnet,deeplabv3+,ccnet,hrnet,transfuse}. However, these methods  usually  acts as backbone for remote sensing image analysis and must combine other network designs to improve the performance.
Fortunately, there are some recent methods dedicated to building footprint extraction.
Yu et al. introduced ConvBNet, which integrates ConvNeXt-XL with a fusion decoder and applies deep supervision during intermediate training stages \cite{yu2023convbnet}.
Besides, some Transformer-based attempts  \cite{xu2023bctnet,transformer_seg01} have been proposed  but did not completely outperform existing CNN-based methods for the reason that  remote dependencies between pixels is not sufficient to
compensate for the lack of a priori knowledge such as local features and location information. Thus CNNs still have an advantage in this field.
Overall,  efficient attention modules, the multi-scale feature propagation and fusion  help to build competitive building extraction solutions \cite{statement0}.

It's worthwhile to note  that the attention mechanism is of benefit  to help feature extraction by selectively focusing on certain information thus allowing neural network learning to be more focused. 
Channel attention \cite{senet} and spatial attention have been shown to be two easy-to-use attention mechanisms in many previous works. 
CBAM (Convolutional Block Attention) further fuses spatial and channel attention \cite{cbam}. SK-Net (Feature-map Attention) is inspired by the nature of cortical neurons \cite{sknet}. ResNeSt (Split Attention, SA)  abstracts the basic architecture of the channel attention and unifies it as a variant of split attention \cite{resnest}. Coordinate Attention (CA) performs separate extraction and fusion of spatial information along vertical and horizontal directions \cite{ca_atten}.

In order to achieve both spatial information retention and better modeling of channel information in a single module for more efficient and targeted semantic feature extraction,
in this paper, we present Split Coordinate Attention (SCA), a novel attention module, which extracts spatial information along vertical and horizontal directions separately and fuses them. We insert it into the ResNet\cite{res_net}  to form an effective SCANet, which obtains SOTA results on two public benchmark dataset. Contributions of this paper include:
\begin{itemize}
  \item We introduce Split Coordinate Attention (SCA), a novel attention module that captures spatially remote interactions using two spatial pooling kernels. Channels are encoded independently, taking into account their placement on both the x and y planes, then splits each feature group separately.
  \item Compared with SA and CA, the parameter amount of our proposed SCA is unchanged or smaller while outperforming both of them.
  \item By inserting into a 2D CNN to form an effective SCANet, it achieves SOTA results on two public building footprint extraction dataset.
\end{itemize}

\section{Split Coordinate Attention Networks}\label{section_scattention}
In this section, we first point the problems of the original Split Attention. We propose Split Coordinate Attention in \ref{SCA}. We introduce SCANet and decoder in \ref{SCANet} and \ref{UNet++}, respectively.

\subsection{Revisit ResNeSt Networks}\label{section_resnest}
As a modification of the ResNet structure, ResNeSt also contains 5 stages, and the feature maps will be downsampled once for each passing stage. If the input  is $512\times512$, then the size of the output  for each stage is $256\times256, 128\times 128, \ldots, 16\times 16$. Adopting this classical structure allows the ResNeSt to be better embedded in the downstream tasks.
\subsubsection{Featuremap Group}
The quantity of feature groups and the quantity of splits within the cardinal group are controlled by two  hyperparameters $K$ and $R$, respectively. As a result, $G = K\times R$ is the  quantity of feature groups.
Each  feature group is transformed by $\left\{\mathcal{F}_{1}, \mathcal{F}_{2}, \ldots, \mathcal{F}_{G}\right\}$
, where  each feature group is depicted  as $U_{i} = \mathcal{F}_{i}(X)$ for $i \in \left\{ 1,2,\ldots,G 
 \right\}$.
\subsubsection{The problem of Split Attention}

According to \cite{senet,sknet}, the collective representation for each cardinality can be derived by merging the sum of elements across various divisions.
$\hat{U}^{k}=\sum_{j=R(k-1)+1}^{R k} U_{j}$ represents the $k$-th cardinal group, where $\hat{U}^{k} \in \mathbb{R}^{H \times W \times C / K}$ for $k \in \left \{ 1, 2,\ldots,K \right \}$. $H,W$ and $C$ represent the sizes of the block output feature maps.

Conventional convolutional layers struggle to capture inter-channel dependencies~\cite{senet}. To address this, Split Attention employs global average pooling across spatial dimensions, effectively gathering comprehensive contextual data that inherently includes channel-specific statistics. The computation for the $c$-th element is expressed as:
\begin{equation}\label{eq:se_extract} 
  s_{c}^{k}=\frac{1}{H \times W} \sum_{i=1}^{H} \sum_{j=1}^{W} \hat{U}_{c}^{k}(i, j) .
\end{equation}

However, global average pooling condenses spatial information globally into channel descriptors, which  leads positional information to be  retained difficultly \cite{ca_atten}.

 \begin{figure}[]
 \centering
    \includegraphics[width=11.9cm]{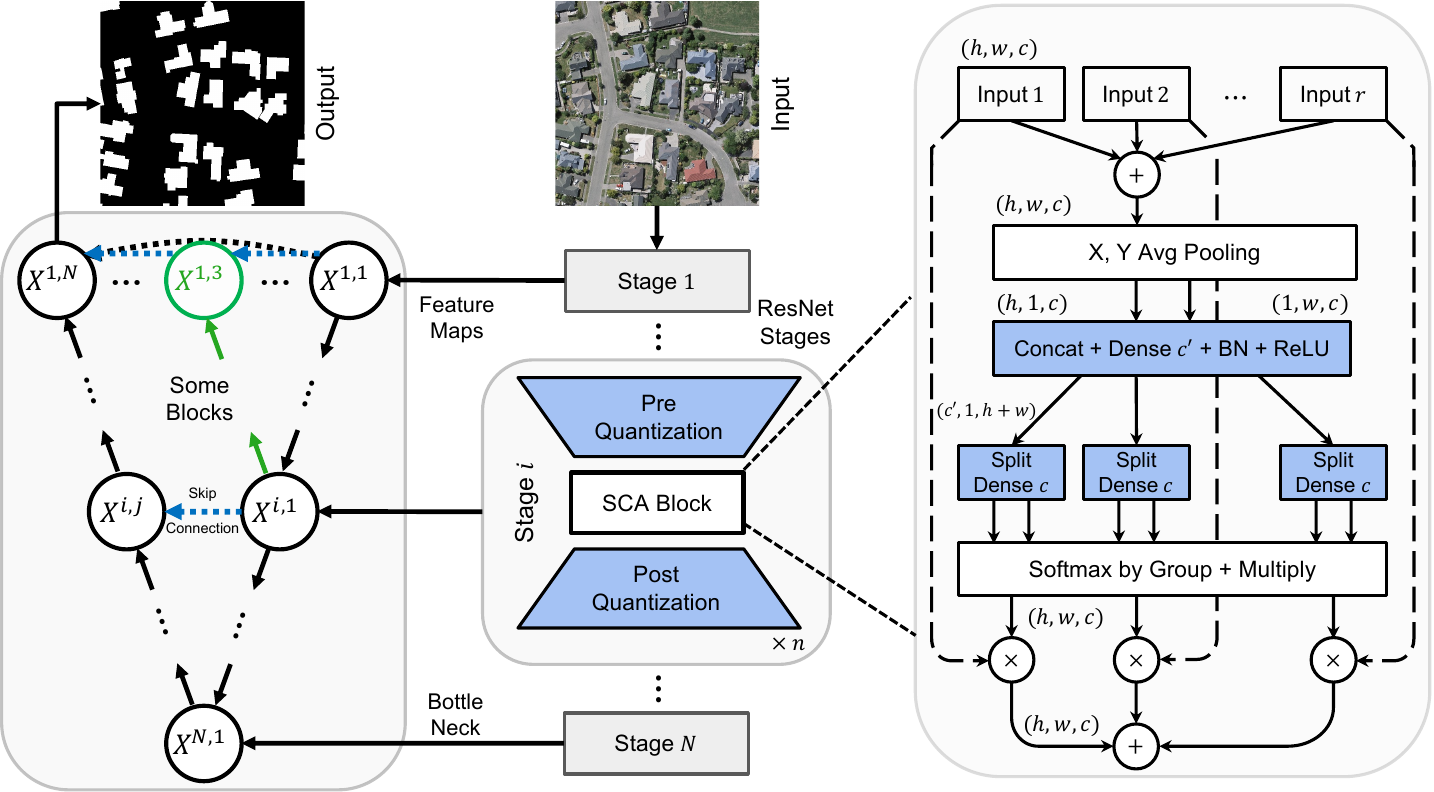}
    \caption{The overall framework of the proposed method. For easy visualization,  $c = C / K$ is set for the figure.}
    \label{fig:overview}
  \end{figure}

\subsection{Split Coordinate Attention}\label{SCA}
To promote attention blocks use precise positional information to capture remote interactions spatially  and to optimize this problem of positional information retention above-mentioned, decomposition of the global pooling described in the Eq.~\ref{eq:se_extract} is done. Fig.~\ref{fig:overview} shows the details of our elaborate Split Coordinate Attention. Overall, to capture spatially remote interactions, by using the two spatial ranges of pooling kernels $(H, 1)$ and $(1, W)$, each channel is encoded on its corresponding horizontal and vertical coordinates.

The $c$-th components of height $ h $ and width $w$  are  respectively determined as:
\begin{equation}
  s_{c}^{h, k}=\frac{1}{W} \sum_{0 \leq j<W} \hat{U}_{c}^{k}(h, j) .
\end{equation}
\begin{equation}
  s_{c}^{w, k}=\frac{1}{H} \sum_{0 \leq i<H} \hat{U}_{c}^{k}(i, w) .
\end{equation}

Thus, the features along two spatial axes are aggregated  using above two transformations, respectively, creating two direction-aware feature maps that encode detailed location information while capturing the global perceptual field. This is quite different from the global average pooling in Split Attention (Eq.\ref{eq:se_extract}). According to the coordinate attention generation designed in \cite{ca_atten}, the above components are combined and represented as:

\begin{equation}
  s_{i}^{k}(c)=\mathcal{G}_i^c\left(\left [s_{c}^{h, k},  s_{c}^{w, k}\right]\right),
\end{equation}
where $\left[\cdot, \cdot\right]$ denotes the process of joining elements along the spatial axis, $s^{k}_i(c) \in \mathbb{R}^{(H + W) \times C / K }$; $\mathcal{G}^c_i$ denotes the activation function applied after convolution, and the result denotes the  $c$-th dimension feature map. $s_{i}^{k}(c)$ is then split along the spatial dimension into two intermediate feature maps, denoted as $t_{i}^{h, k}(c)\in \mathbb{R}^{H \times C / K }$ and $t_{i}^{w, k}(c)\in \mathbb{R}^{W \times C / K }$. 
To form dense connections, let $\mathcal{S}^h_c$ and $\mathcal{S}^w_c$ represent two Dense connections to transform the quantity of channels of $t_{c}^{h, k}$ and $t_{c}^{w, k}$ so as to be unified with $\hat{U}_{c}^{k}$:

\begin{equation}
  \begin{cases} 
    u_{i}^{h, k}(c) & =\sigma\left[\mathcal{S}^h_c\left(t_{i}^{h, k}(c)\right)\right], \\ 
    u_{i}^{w, k}(c) & =\sigma\left[\mathcal{S}^w_c\left(t_{i}^{w, k}(c)\right)\right],
 \end{cases}
\end{equation}
where $\sigma$ is the sigmoid function. These dense connections allow for better information flow and feature reuse across different spatial dimensions.

The weighted combination on splits is utilized to generate each feature map channel. 
By utilizing the channel soft attention, the weighted fusion of the cardinal group $V^k \in \mathbb{R}^{H \times W \times C / K }$  is aggregated.
The $c$-th channel is calculated as follows:

\begin{equation}
  V_{c}^{k}=\hat{U}^k_c\sum\limits_{i = 1} ^ R {u_{i}^{h, k}(c)\times u_{i}^{w, k}(c)}.
\end{equation}

This formulation allows for the integration of both horizontal and vertical attention weights, enabling the network to focus on relevant spatial locations more effectively.

\subsection{SCANet Block} \label{SCANet}
The cardinal group is represented as concatenating $V^i$ across the channel dimension, i.e.,
$V = Concat\{V^1, V^2, \ldots, V^K \}$.
Consistent with the standard residuals module, our final output $Y$ is also concatenated using shortcut $Y = V + \mathcal{T}(X)$,
where $\mathcal{T}$ denotes a transformation that is used to align the tensor's shape.

\subsection{UNet++ Decoder} \label{UNet++}
The features obtained by a encoder with  high-dimensional abstract modeling  need a powerful decoder to parse information. In this paper, we choose UNet++~\cite{unet++} as decoder to parse the highly abstract feature maps provided by SCANet.

\begin{figure}[t]
    \centering
    \begin{subfigure}[b]{5.5cm}
    \centering
      \includegraphics[width=\linewidth]{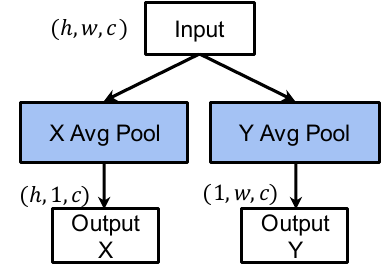}
      \caption{X, Y Avg Pooling}
      \label{fig:xyavgpool}
    \end{subfigure}
    \begin{subfigure}[b]{5.5cm}
    \centering
      \includegraphics[width=\linewidth]{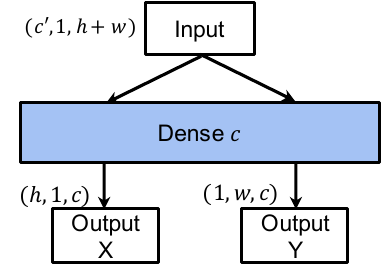}
      \caption{Split Dense}
      \label{fig:splitdense}
    \end{subfigure}
    \caption{The figure shows the X,Y Avg Pooling module (a) and the Split Dense module (b) of the proposed Split Coordinate Attention.}
\end{figure}
  

\section{Experiments and Results}
\subsection{Benchmark Datasets}
Two benchmark datasets, the MASS Dataset~\cite{mass} and the WHU Dataset~\cite{whu}, are used to validate our SCANet. Each WHU tile is 512 × 512 pixels, with 4736 tiles for training, 1036 for validation, and 2416 for testing. Each image in the MASS measures $1500 \times 1500$. The dataset consists of 137 training images, 4 validation images, and 10 test images.  This dataset has a lower ground resolution and labeling accuracy compared to the WHU Building Dataset.For both datasets above, we uniformly crop into $512 \times 512$  for the experiments.

\subsection{Training Implementation}
This paper employs consistent hyperparameter settings across all experiments. For the WHU and MASS datasets focusing on buildings, batch sizes of 16 and 4 were used, respectively. The loss function combines Binary Cross Entropy and Dice Loss, and the optimizer used is AdamW~\cite{adamw} with an initial learning rate of 1e-3. The validation set's intersection over union (IoU) is evaluated after each epoch. If the IoU does not improve after 10 epochs, the learning rate is halved. The model is implemented using PyTorch 1.13 on an Ubuntu 20.04 system with CUDA version 11.8 and an RTX3090 GPU.

\subsection{Comparison With SOTA Methods}

We quantitatively compare different SOTA methods with our SCANet on this two public dataset, including universal segmentation methods, i.e., PSPNet, DeeplabV3+, HRNet, CCNet and TransFuse-L, and excellent dedicated building footprint extraction methods, such as BCTNet~\cite{xu2023bctnet} and ConvBNet~\cite{yu2023convbnet}. Besides, we also combine ResNeSt and UNet++ for comparison with our method.

To ensure the fairness of the experiments, HRNetV2-W48 is selected in HRNet. ResNet-101 is selected as the backbone  in PSPNet, Deeplabv3+, CCNet and ResNeSt . ResNet50 and DeiT-Base are selected as the backbone of TransFuse-L.

Table \ref{tab:whu_table} and Table \ref{tab:mass_table} present the quantitative results, where bold indicating the best results. Compared with all other methods, our proposed SCANet has more significant performance on the two benchmark datasets. The model demonstrates superior performance, attaining peak values for Intersection over Union, Precision, and F1-Score in both datasets.
Particularly compared with the SOTA methods, SCANet improves the IoU by $0.46\%$ on the WHU Building Dataset and $0.45\%$ on the Massachusetts Building Dataset. 

\begin{table}[!ht]
  \centering\caption{The quantitative results of the SOTA Methods and ours on the WHU  Building Dataset (\%)}\label{tab:whu_table}
 \begin{tabular}{c|cccc}
  \hline
  \toprule
      \textbf{Methods} & \textbf{IoU↑} & \textbf{Precision↑} & \textbf{Recall↑} & \textbf{F1-Score↑} \\ \midrule
        PSPNet & 89.08  & 94.46  & 93.99  & 94.22  \\ 
        DeeplabV3+ & 88.52  & 92.92  & 94.93  & 93.91  \\ 
        CCNet & 89.33  & 94.87  & 93.86  & 94.36  \\ 
        HRNet & 90.35  & 95.21  & 94.65  & 94.93  \\ 
        TransFuse-L & 90.10  & 94.41  & 95.18  & 94.79  \\ 
        ResNeSt \& UNet++ & 90.51  & 95.42  & 94.63  & 95.02  \\
        ConvBNet & 91.12  & 95.17  & 95.65  & 95.11  \\ 
        BCTNet & 91.15  & 95.47  & 95.27  & 95.37  \\  \midrule
        SCANet (Ours) & \textbf{91.61} & \textbf{95.92} & \textbf{95.67} & \textbf{95.79} \\  \bottomrule
  \end{tabular}
\end{table}

\begin{table}[!ht]
  \centering\caption{The quantitative results of the SOTA Methods and ours  on the Massachusetts Building Dataset (\%)}\label{tab:mass_table}
\begin{tabular}{c|cccc}
  \hline
  \toprule
      \textbf{Methods} & \textbf{IoU↑} & \textbf{Precision↑} & \textbf{Recall↑} & \textbf{F1-Score↑} \\ \midrule
      PSPNet & 69.43  & 84.02  & 79.99  & 81.95  \\ 
        DeeplabV3+ & 73.00  & 86.24  & 82.63  & 84.39  \\ 
        CCNet & 69.81  & 84.75  & 79.84  & 82.22  \\ 
        HRNet & 73.97  & 87.40  & 82.79  & 85.03  \\ 
        TransFuse-L & 72.02  & 82.54  & \textbf{84.97}  & 83.74  \\ 
        ResNeSt \& UNet++ & 74.43  & 87.44  & 83.35  & 85.34  \\
        BCTNet & 75.04  & 87.57  & 83.99  & 85.74  \\ \midrule
        SCANet (Ours) & \textbf{75.49}  & \textbf{88.38}  & 84.30  & \textbf{86.29} \\ \bottomrule
  \end{tabular}
\end{table}

\begin{figure}
\centering
\includegraphics[width=11.9cm]{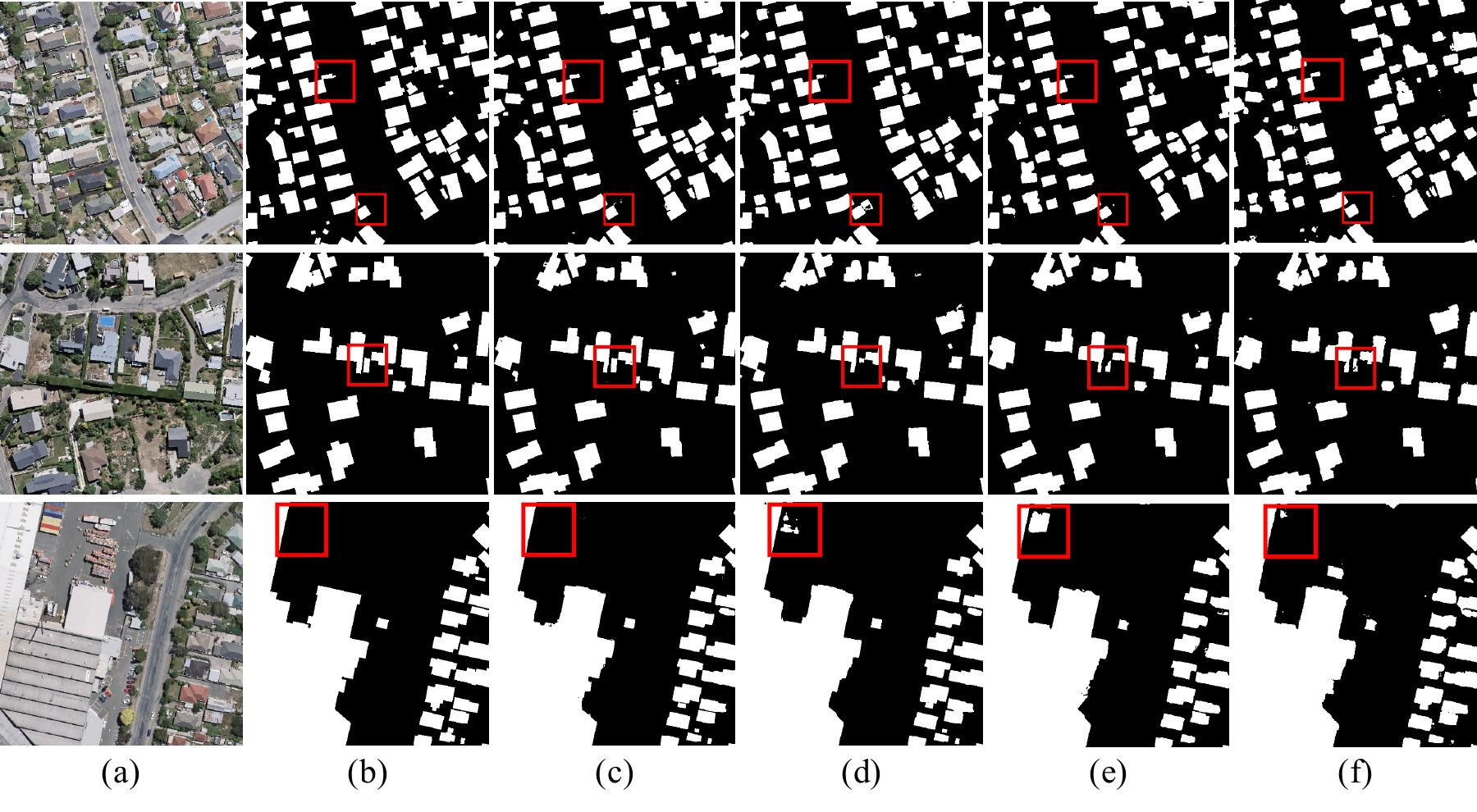}
\caption{Qualitative results on the WHU Dataset. (a) Image. (b) GT. (c) SCANet. (d) ResNeSt \& UNet++. (e) HRNet. (f) ConvBNet.}
\label{fig:whu_vis}
\end{figure}

\begin{figure}
  \centering
 \includegraphics[width=11.9cm]{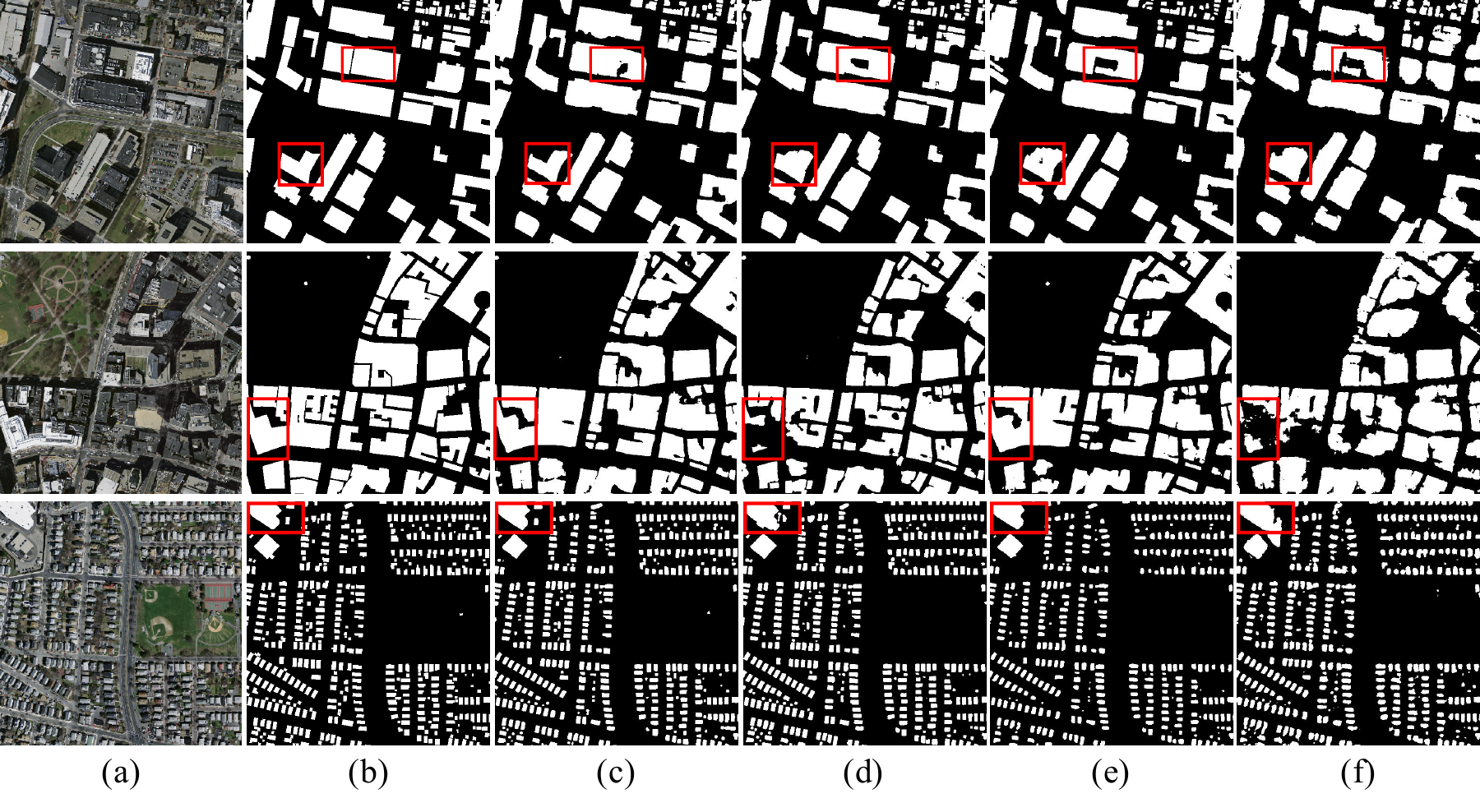}
  \caption{Qualitative results on the MASS Dataset. (a) Image. (b) GT. (c) SCANet. (d) ResNeSt \& UNet++. (e) HRNet. (f) ConvBNet.}
  \label{fig:mass_vis}
  \end{figure}

Besides,  we present some qualitative results of the two datasets in Figure \ref{fig:whu_vis} and Figure \ref{fig:mass_vis} to further evaluate our SCANet, respectively. For WHU Dataset, we choose three representative building scenes, i.e., large-area buildings, sparse medium-area buildings and dense small-area buildings. As shown in Figure \ref{fig:whu_vis}, in the first scene, our method has the best performance and can distinguish the background similar to the buildings more cleanly. In the second scene, for smaller building edges, our method is able to capture the details better and can segment them more completely. While ConvBNet can extract details effectively, its segmentation results lack smoothness. Our proposed method excels in addressing this issue, particularly in the third scene, by accurately focusing on small, irregular buildings and correctly distinguishing similar backgrounds. In conclusion, our method obtains the best segmentation results on this dataset.

Similar to WHU Dataset, we also choose three representative building scenes from Massachusetts Dataset, including dense small buildings, continuous large area buildings and scattered large area buildings. In the first scene, SCANet is able to correctly capture the details of the small buildings next to the large-area buildings and give accurate segmentation results, while the other methods do not distinguish them well. In the second scene, our method works best for the hollow part in the middle of the building, which can be more completely separated from the building. In the third scene, our method is able to segment the shadow part of the irregular dark building better, and we note that HRNet is more effective in segmenting the hollow part of the building due to the high resolution of the feature map, but it is less able to extract the semantic information, which leads to the wrong segmentation of the building into hollow parts. Overall, our method is able to better capture the edge details of the building, and at the same time can better distinguish the building from the background.

\subsection{Ablation Study}
\begin{table}[]
\centering\caption{
The experiment results of SA, CA and SCA on the WHU Building Dataset (\%)
}\label{tab:ablation}
\begin{tabular}{@{}l|c|c|cc@{}}
\toprule
\textbf{\ Settings} & \textbf{Decoder} & \textbf{Params (M)↓} & \textbf{IoU↑} & \textbf{F1-Score↑} \\ \midrule
\ \ Baseline & UNet++ & 68.0 & 88.21 & 93.63 \\
\ \ +SA      & UNet++ & 73.2 & 90.51 & 95.02 \\
\ \ +CA      & UNet++ & 79.2 & 90.62 & 95.06 \\
\ \ +SCA     & UNet   & 55.3 & 90.41 & 94.91 \\
\ \ +SCA     & UNet++ & 73.2 & \textbf{91.61} & \textbf{95.79} \\ \bottomrule
\end{tabular}
\end{table}
Table~\ref{tab:ablation} 
summaries the experiment results of SA, CA and SCA using the same baseline on the WHU Building Dataset. We also compare UNet++ and UNet as the decoders to discuss the influence of them when  parsing
the highly abstract feature maps provided by SCANet. When UNet++ is used as the decoder, our method achieves the best results and has less or the same number of parameters compared with CA and SA.

\subsection{Complexity Analysis}
To comprehensively compare the performance and parameter count of SCANet with other methods, we have listed the backbone networks used by different methods, their corresponding model parameter counts, and IoU metrics on two benchmark datasets in Tab~\ref{tab:params}. By analyzing the data, we can see that SCANet consistently shows stable performance improvements regardless of whether ResNet~\cite{res_net} or ResNeSt~\cite{resnest} is used as the backbone.
\begin{table}[]
\setlength\tabcolsep{5pt}
\centering
\caption{Comparison of parameters of each method.}\label{tab:params}
\begin{tabular}{c|c|c|ccc}
\toprule
\multirow{2}{*}{\textbf{Methods}} & \multirow{2}{*}{\textbf{Backbone}} & \multirow{2}{*}{\textbf{Params(M)↓}} & \multicolumn{2}{c}{\textbf{IoU↑}}    \\ 
                               &                      &       & \textbf{WHU}    & \textbf{MASS}   \\ \midrule
BCTNet                         & ResNet-18 + PVTv2-B2 & 78.3  & 0.9115          & 0.7504          \\
Swin UNETR                     & -                    & 57.1  & 0.8766          & 0.7028          \\
Swin-B                         & -                    & 121.2 & 0.9004          & 0.7392          \\ \midrule
\multirow{4}{*}{ResNet + CA}   & ResNet-14            & 36.3  & 0.8742          & 0.7167          \\
                               & ResNet-26            & 45    & 0.8813          & 0.7203          \\
                               & ResNet-50            & 59.1  & 0.8821          & 0.7261          \\
                               & ResNet-101           & 88.7  & 0.8859          & 0.7291          \\ \midrule
\multirow{4}{*}{ResNeSt}       & ResNet-14            & 34    & 0.8946          & 0.7366          \\
                               & ResNet-26            & 40.5  & 0.8988          & 0.7432          \\
                               & ResNet-50            & 50.9  & 0.9006          & 0.7445          \\
                               & ResNet-101           & 73.2  & 0.9011          & 0.7449          \\ \midrule
\multirow{4}{*}{SCANet (ours)} & ResNet-14            & 34    & 0.9061          & 0.7435          \\
                               & ResNet-26            & 40.5  & 0.9085          & 0.7488          \\
                               & ResNet-50            & 50.9  & 0.9152          & 0.7508          \\
                               & ResNet-101           & 73.2  & \textbf{0.9161} & \textbf{0.7549} \\ \bottomrule
\end{tabular}
\end{table}

\subsection{Feature Visualization}
\begin{figure}
\begin{center}
\includegraphics[width=11.9cm]{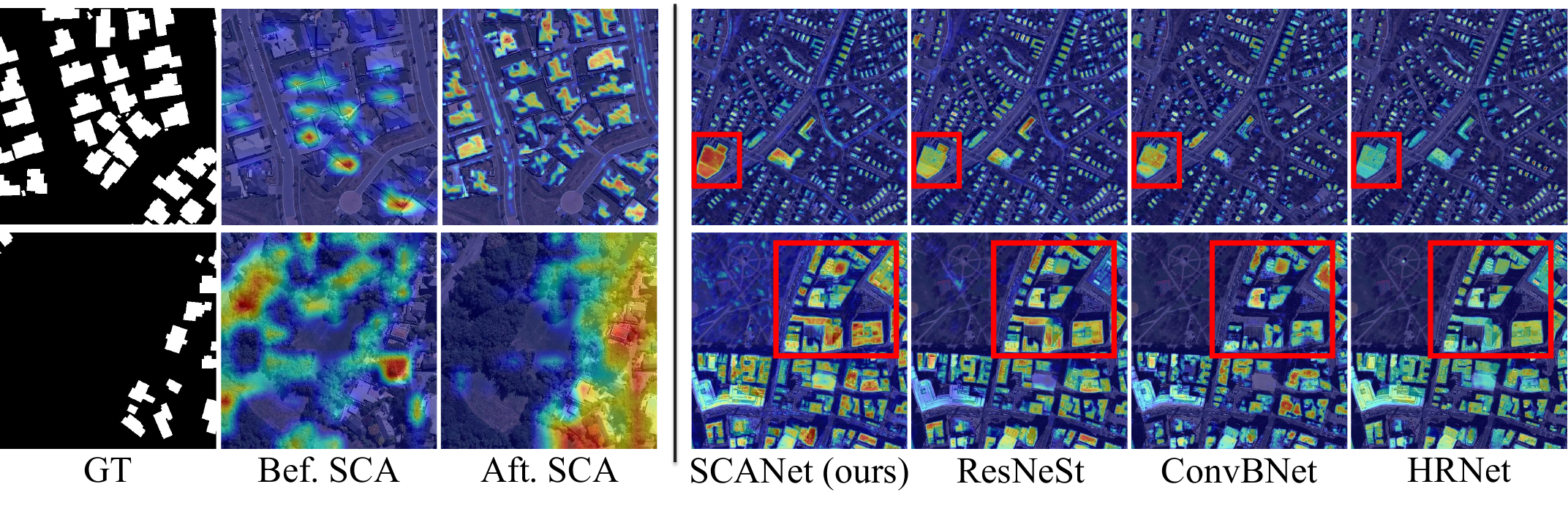} \caption{Grad-CAM++ visualization of the proposed SCANet and other methods on the Massachusetts Building Dataset.} \label{fig:gradcam}
\end{center}
\end{figure}
To further explain the capacity of our model to capture semantic information, we visualize some feature maps  of the networks.

Gradient-weighted class activation mapping (Grad-CAM) is used to qualitatively interpret the performance of deep learning based methods, which use mask's gradient information to generate heat maps that show the regions of interest and the level of attention that the model pays when generating segmented images.  We use Grad-CAM++~\cite{gradcampp} to visualize the feature maps as RGB images in Figure \ref{fig:gradcam}. 

As shown in Figure \ref{fig:gradcam}, the proposed SCANet is able to focus more attention on the building compared to other methods, which helps to improve the edge details of the building footprint extraction. ResNeSt, however, pays less attention to buildings. ConvBNet is able to achieve better attention for large buildings, but has insufficient attention for small buildings. HRNet has less attention  for whether large or small  buildings. The above analysis qualitatively proves that our proposed Split Coordinate Attention is effective. We also visualize the attention distribution of the same feature before and after being processed by SCA. It is evident that SCA can capture and localize long-range interactions, thereby achieving better performance.

\begin{figure}
    \centering
    \includegraphics[width=10cm]{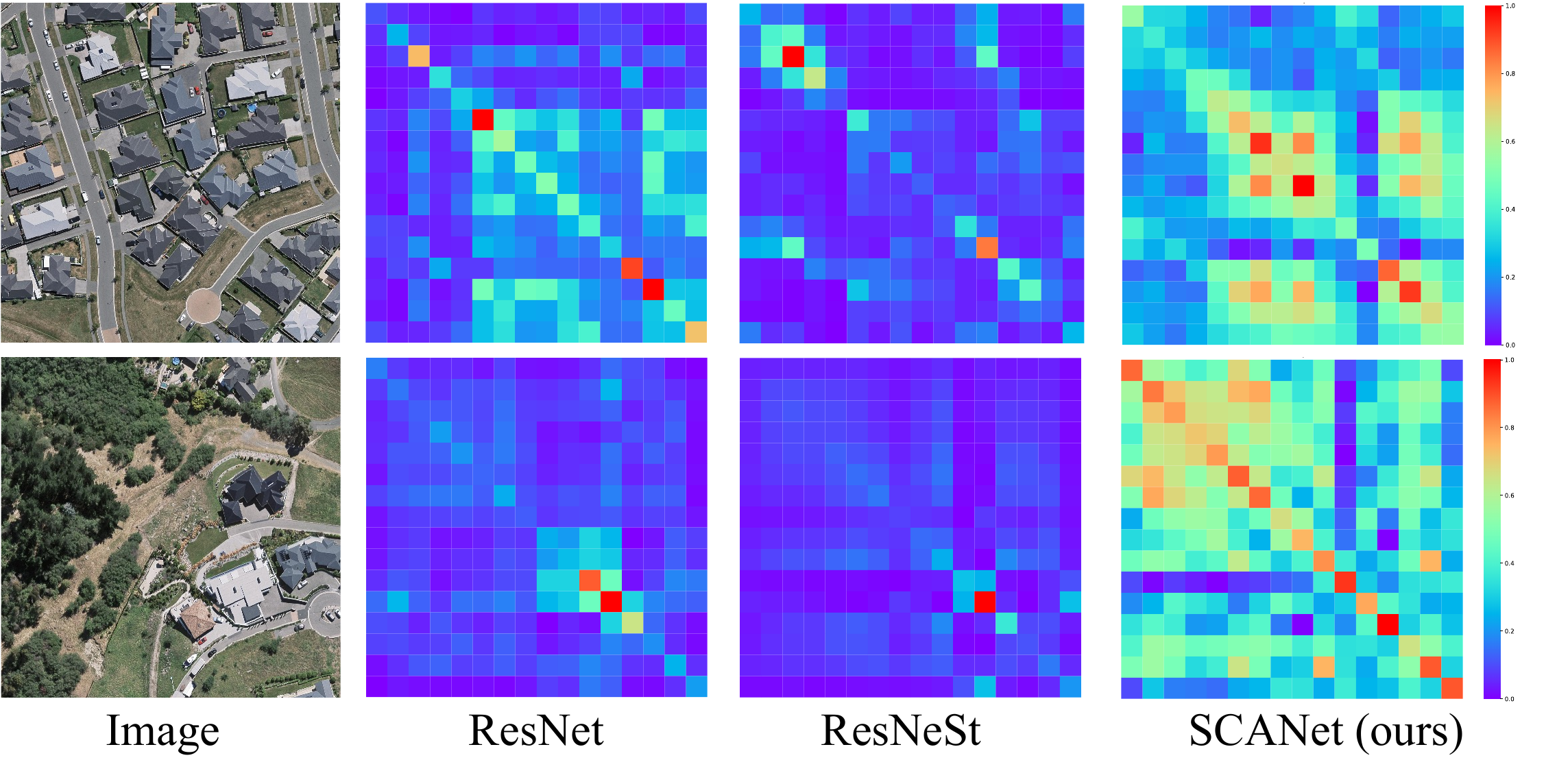}
    \caption{The diagonal similarity of different encoders in the final stage.}
    \label{fig:sim_matrix}
\end{figure}

To test the effectiveness of the designed module, we present the visualization results of the diagonal pixel similarity matrix for different Encoders at the final stage (Stage-4) in Figure \ref{fig:sim_matrix}. This visualization approach is referred from previous studies \cite{xie2021segformer,emo} used to assess the receptive field size of models. From the results shown in Figure \ref{fig:sim_matrix}, we can conclude that when using ResNet and ResNeSt as the base encoders, the extracted features mainly exhibit short-range correlations in adjacent areas. However, with our SCANet, long-range spatial correlations are significantly improved. Our method demonstrates a larger effective receptive field, as evidenced by enhanced feature similarity between distant locations.

\section{Discussion}
The Spatial Context Attention (SCA) module, as a CNN-based plug-and-play component, offers significant advantages in image segmentation. While testing SCA with a wider variety of backbone networks, including transformer-based models, could be considered, it's important to highlight the strengths of our current approach.

Our CNN-based module provides notable benefits in computational efficiency and ease of integration, crucial factors in remote sensing image analysis where resources may be limited. Despite its simplicity, SCA demonstrates remarkable effectiveness. Our comparative analysis already includes a range of baseline networks, including transformer-based models like TransFuse, ConvBNet, BCTNet, Swin UNETR, and Swin-B. Notably, our ResNet-based integration consistently outperforms these more complex models in most cases.This performance superiority validates the robustness and versatility of the SCA module, challenging the notion that increased model complexity necessarily leads to better results. While future research could explore SCA's compatibility with other architectures, the current results strongly support the effectiveness of our CNN-based approach.

In conclusion, the SCA module not only introduces an effective approach for remote sensing image segmentation but also demonstrates how targeted improvements to CNN architectures can yield significant advancements, offering a balanced solution that combines high performance with practical applicability.

\section{Conclusion}
This paper present a novel plug-and-play attention module  Split Coordinate Attention (SCA), which can be inserted into 2D CNN to form an effective SCANet, achieving SOTA performance on two public building footprint extraction dataset, such as WHU Building Dataset and Massachusetts Building Dataset.

\begin{credits}
\subsubsection{\ackname} This work is supported in part by the Youth Science Foundation of Guangxi Natural Science Foundation (Grant No.2023GXNSFBA026018), the Project of Improving the Basic Scientific Research Ability of Young and Middle-Aged Teachers in Universities of Guangxi Province (Grant No.2023KY0223), the National Natural Science Foundation of China (Grant No.62076077),  the Guangxi Science and Technology Major Project (Grant No.AA22068057).

\end{credits}
%
%
%
\bibliographystyle{splncs04}
\bibliography{0993}
%




\end{document}